\documentclass[sigconf, screen, table]{acmart}
\renewcommand\footnotetextcopyrightpermission[1]{} 
\usepackage{tikz}
\usepackage{collcell}
\usepackage{etoolbox}
\usepackage{xstring}
\usepackage{pgf}
\usepackage{xurl}
\usepackage{stfloats}

\AtBeginDocument{
  \providecommand\BibTeX{{
    \normalfont B\kern-0.5em{\scshape i\kern-0.25em b}\kern-0.8em\TeX}}}

\begin{document}

\title{Comparing PCG metrics with Human Evaluation in Minecraft Settlement Generation}

\author{Jean-Baptiste Hervé}
\affiliation{
  \institution{University of Hertfordshire}
  \streetaddress{College Lane}
  \city{Hatfield}
  \country{UK}}
\email{jbaptiste.herve@gmail.com}

\author{Christoph Salge}
\affiliation{%
  \institution{University of Hertfordshire}
  \streetaddress{College Lane}
  \city{Hatfield}
  \country{UK}}
\email{ChristophSalge@gmail.com}

\newcommand*{\MinNumber}{-1.0}%
\newcommand*{\MidNumber}{0.0} %
\newcommand*{\MaxNumber}{1.0}%

\newcommand{\ApplyGradient}[1]{%
 \IfDecimal{#1}{
   \ifdim #1 pt > \MidNumber pt
        \pgfmathsetmacro{\PercentColor}{max(min(100.0*(#1 - \MidNumber)/(\MaxNumber-\MidNumber),100.0),0.00)} %
        \edef\x{\noexpand\cellcolor{green!\PercentColor!white}}\x\textcolor{black}{#1}
   \else
       \pgfmathsetmacro{\PercentColor}{max(min(100.0*(\MidNumber - #1)/(\MidNumber-\MinNumber),100.0),0.00)}
       \edef\x{\noexpand\cellcolor{red!\PercentColor!white}}\x\textcolor{black}{#1}
   \fi
  }{#1}
}

\newcommand{\ApplyBoldGradient}[1]{%
 \IfDecimal{#1}{
   \ifdim #1 pt > \MidNumber pt
        \pgfmathsetmacro{\PercentColor}{max(min(100.0*(#1 - \MidNumber)/(\MaxNumber-\MidNumber),100.0),0.00)} %
        \ifdim #1 pt > 0.77 pt
            \edef\x{\noexpand\cellcolor{green!\PercentColor!white}}\x\textcolor{black}\x\textbf{#1}%
        \else
            \edef\x{\noexpand\cellcolor{green!\PercentColor!white}}\x\textcolor{black}{#1}%
        \fi
   \else
       \pgfmathsetmacro{\PercentColor}{max(min(100.0*(\MidNumber - #1)/(\MidNumber-\MinNumber),100.0),0.00)}
        \ifdim #1 pt < -0.77 pt
            \edef\x{\noexpand\cellcolor{red!\PercentColor!white}}\x\textcolor{black}\x\textbf{#1}%
        \else
            \edef\x{\noexpand\cellcolor{red!\PercentColor!white}}\x\textcolor{black}{#1}%
        \fi
   \fi
  }{#1}
}

\newcolumntype{R}{>{\collectcell\ApplyGradient}{c}<{\endcollectcell}}
\newcolumntype{S}{>{\collectcell\ApplyBoldGradient}{c}<{\endcollectcell}}

\renewcommand{\shortauthors}{Hervé and Salge}

\begin{abstract}
There are a range of metrics that can be applied to the artifacts produced by procedural content generation, and several of them come with qualitative claims. In this paper, we adapt a range of existing PCG metrics to generated Minecraft settlements, develop a few new metrics inspired by PCG literature, and compare the resulting measurements to existing human evaluations. The aim is to analyze how those metrics capture human evaluation scores in different categories, how the metrics generalize to another game domain, and how metrics deal with more complex artifacts. We provide an exploratory look at a variety of metrics and provide an information gain and several correlation analyses. We found some relationships between human scores and metrics counting specific elements, measuring the diversity of blocks and measuring the presence of crafting materials for the present complex blocks.  
\end{abstract}


\begin{teaserfigure}
  \includegraphics[width=\textwidth]{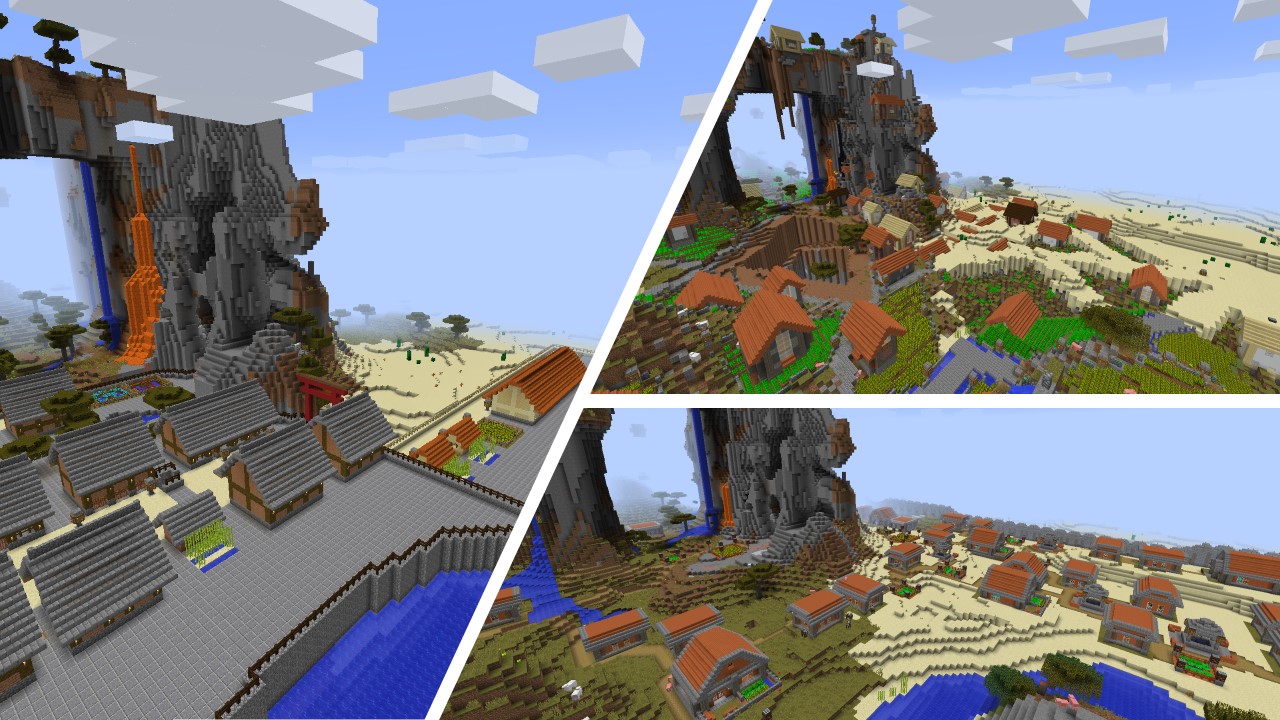}
  \caption{Several computer generated settlements in a Minecraft world, taken from the three best scoring entries in the 2020 GDMC Settlement Generation Challenge.}
  \Description{Minecraft Image}
  \label{fig:teaser}
\end{teaserfigure}

\maketitle
\pagestyle{plain}

\section{Introduction}

Procedural content generation (PCG) is a part of Game AI that aims to create game content algorithmically. In the past it has been used to generate game levels, such as race tracks \cite{togelius2007towards} or strategy maps \cite{togelius2010towards}, game assets, such as trees \cite{speedtree}, or even full sets of game rules \cite{TogeliusAutomatic,font2013card,Khalifa2017,mahlmann2013modelling,mahlmann2012evolving}. It can be used by itself, but also to critique, suggest or refine game content. One recurring challenge in procedural content generation (PCG) is evaluating the output \cite{summerville2017understanding}. The core constraint is usually playability - aimed to ensure that a game can actually be used as such. One common approach here is to use AI gameplay \cite{silva2018exploring} to test that a game can be won, or to determine basic imbalances \cite{salge2008using,Volz2016}. Since generating content is usually fast and cheap \cite{smith2017we}, the aim is usually to test content automatically - but other approaches with a human in the loop exist. Beyond playability there is also the aim to produce ``good'' content, which is usually tied to evoking specific forms of human experience, such as fun or engagement, or to just get a better, quantifiable understanding of what is being produced \cite{canossa2015towards}. To this end, several PCG metrics have been developed over the years, particular to be used within a generate-and-test or search based PCG framework \cite{Togelius2011}. When we talk about metrics in this paper, we mean any kind of algorithm that can be applied to generated content and produce a quantitative result. 

Initial approaches here were focused on the \emph{Expressive Range} of a generator \cite{smith2010analyzing}, using metrics to quantify the diversity of a given generator. A different output from the metric being proof that the generated artifact was different. When it comes to evaluating the qualitative experience evoked by an artifact, research tends to use metrics that try to capture specific properties of a piece of content, like its difficulty or its aesthetic. While most of these metrics have been developed and used in different publications, they are heavily defined by the context (most of the time a specific game genre) in which they have be designed, and with a particular objective in mind. In particular, there has been a great focus on level generation for the game Super Mario and similar platformers \cite{snodgrass2017procedural,shaker2012evolving,yannakakis2011experience,dahlskog2012patterns}. We wonder if those metrics can be generalized and will still be relevant in a different context.

For this paper we are interested to see if some of those metrics can be applied to Minecraft settlements. This was prompted by our interest in the GDMC AI settlement generation competition \cite{salge2018generative}  - an AI challenge were participants build algorithms that can make ``interesting'' settlements adapted to unknown landscapes. The GDMC challenge was specifically designed as a PCG competition, to foster interest in both adaptive and holistic PCG. The term holistic here refers to the idea that a single artifacts has to fulfill several criteria that cannot be easily separated. While a PCG level and PCG music should be orchestrated \cite{antoniosliapis2015} to fit with each other, they can still be generated as separate artifacts. A settlement and its buildings in contrast need to be functional, aesthetic, adaptive, etc, all at the same time. A Minecraft map is overall also a rather large design space, so there is a concern that any metric is only capturing a partial projection of the complex artifact - hindering a clear divide and conquer approach to content evaluation. Currently the GDMC uses exclusively human judges for its evaluation step. We are interested to see if we can adapt some of the existing metrics, or develop new ones, suitable for the Minecraft map domain.  

This leads us to our first main research question for this paper: How well do platformer metrics generalize to other domains - specifically the domain of Minecraft settlements?
To attempt to answer this, we will report on our effort to implement existing metrics in the Minecraft domain, porting mostly 2d-tile based algorithms to 3d-block based implementations. We first surveyed the literature in search of suitable candidates which we could directly apply to the Minecraft settlements. But during this process we learned that all metrics required some degree of adaptation, either to provide some domain specific information, deal with the shift to 3d, or to address some other problems that arise for this specific application. Several design choices needed to be made, and while some were more principled than others, we tried to document this process to outline, as an example, what it would take to make this kind of adaptation. In this process we also developed several new metrics for the 3d block domain.

This leads us to the second main question - are those existing or newly developed metrics related to human experience or judgment? While several of the existing metrics for platformers have acquired qualitative claims over time, a lot of those have been based on common sense assumptions. There have been few works that specifically tried to ground existing PCG metrics to human experience \cite{marino2015empirical}, and none so far for the Minecraft domain - a gap we aim to close. Grounded metrics can then even be used to generate specific levels geared towards certain experiences, or following the preferences of individual players~\cite{yannakakis2011experience}. One of the reasons we chose the GDMC settlement generation challenge as a test bed was the fact that there are both generators and a published set of human evaluation scores for each generator - which allows us to compare the metrics for various submitted settlements generators with the human assigned scores for them from the competition. As far as we know, this is one of the most complex and messy PCG artifacts to date to be analyzed with automatic metrics, so we were also interested in how robust a metric based approach is to complex content generation?

\subsection{Overview}

The paper is structured as follows. We first give some more background on the GDMC competition, with particular focus on the human evaluation criteria and the process of human evaluation. We also give some introduction to the terms and concepts of a Minecraft settlement, to give context to the artifacts analyzed here. We then discuss the metrics we used in this paper, giving a sketch of their origin, purpose and implementation here. Following this we provide a statistical analysis of the results. We discuss both the results, and their wider implications. 

\section{Background}

\subsection{Minecraft Settlement}

Minecraft \cite{game:Minecraft} is a voxel-based, open world game where the player runs around in a 3d blockworld, interacting with the world by picking up blocks, placing new blocks, and making various items and new blocks. It has been overwhelmingly popular, and has been compared to LEGO on a computer \cite{thompson2016minecraft,overby2015virtual}. The LEGO comparison is not just due to the block-based nature of Minecraft, but also a reference to its open-endedness. While the game has traditional game mechanics such as combat, death and winning the game by slaying a Dragon, many players ignore those aspects, or turn them off, and enjoy Minecraft as a creative Sandbox. 
A Minecraft Settlement is a series of structures arranged to look like a village or town. There are both game-generated and player-generated settlements \cite{nicolasvalencia2016}. They do provide game-relevant functions and affordances. This includes a protection from dangerous mobs, access to blocks needed to craft tools or grow food, and a place to store items. While these functional elements mirror their functionality of a real world settlement to some extent there is a great degree of variety of how those ``settlements'' can look, and how far they deviate from the classical fantasy village look of the settlements generated by Minecraft itself. This is possible because many of the constraints of regular settlements, such as physics or a need for sanitation or warmth, are simply not present in the game world. Nevertheless, many players still spend time and effort to built up the decorative aspects of settlements \cite{nicolasvalencia2016} for their personal pleasure, to tell a story, or for other reasons. A settlement can be inspired by a specific place or time period, a specific cultural reference, or even be designed from scratch. Settlements are a popular type of structure to build, and it is common among the players to showcase their creations \cite{nicolasvalencia2016}. 

Minecraft settlements are built by arranging blocks, the basic unit of the Minecraft world. A block is a cube, with a simple texture, that is located at a specific, discrete 3d coordinate. Every 3d coordinate contains exactly one block, such as sand, a chest, or lava, which gives the game world a coarse-grained regularity. Many of the blocks can be found in the open world as resources. They can then be refined, or combined in order to craft new types of blocks. In many ways the blocks are the comparable unit to the tile in platformers.

We will see that the generators we look at later will interact with the game map, which contains the settlement just as well as the terrain, as they are both just different arrangements of the basic blocks. The generators can read blocks and place blocks (represented simply by integers to denote their type) at discrete 3d coordinates. This abstracts away some of the game elements of Minecraft, such as mobs, i.e. mobile entities such as cows or lit dynamite. See Fig.~\ref{fig:teaser} for three examples of settlements generated by algorithms submitted to the GDMC challenge. 

While Minecraft has a range of interesting game mechanics, we did not introduce them here in detail, as we limited our search for metrics to those that could be applied to the block representation without the need to simulate the actual agent-based gameplay. Note that the evaluation done by humans, on the other hand, was performed by human judges using the actual game to walk through and interact with the settlements (in creative mode). 

\subsection{GDMC Settlement Generation Competition}

The Generative Design in Minecraft Competition (GDMC) is a PCG competition in which competitors submit a settlement generator \cite{salge2018generative}. These generators work by adding and removing Minecraft blocks. All the submitted generators are then tested on 3 maps with a fixed size of 256x256, which are selected by the organizers \cite{salge2020ai}. All the generated settlements are then stored and sent to the jury. The jury is composed of a permanent board and guests. This jury includes experts in various field, such as AI, Game Design or Urbanism. However, due to various reasons, only a subset return a complete evaluation.  Each judge scores the settlements between 0 to 10 points, in each of the following categories : \emph{Adaptability, Functionality, Narrative, Aesthetic}. \emph{Adaptability} is how well the settlement is adapted to its location - how well it adapts to the terrain, both on a large and small scale. \emph{Functionality} is about what affordances the settlement provides, both to the Minecraft player and the simulated villagers. It covers various aspects, such as food, production, navigability, security, etc. \emph{Narrative} reflects how well the settlement \emph{itself} tells an evocative story about its own history, and about who its inhabitants are (There is a separate bonus challenge about also adding a written PCG text that tells the story of the settlement\cite{Salge2019}). \emph{Aesthetic} is a rating of the overall look of the settlements. In the competition, the rating of each category is computed for each generator by averaging (mean) across all judge's scores. The rating works in the following way : a grade of 5 means that the result looks human made, a 6-9 correspond to what we would expect from an expert human, and finally a 10 would be attributed to a ``superhuman performance''. The judges provide for each generator, after looking at three maps, one score for each of the four categories.
The overall score of the generators is then obtained by a mean average over the four categories. For details, see the instructions given to the judges in the appendix~\ref{ref:instructions}. The number of judges that returned a verdict was 8 for 2018, 11 for 2019 and 8 for 2020.

The human data we are working with is the average scores for the generator, which we compute by averaging over the scores given by all the judges in the year it was submitted. Therefore, for each generator we have 5 scores: the overall score, and one for each of the categories, adaptivity, functionality, narrative and aesthetics.

\section{Metrics}
In this section we will introduce and describe the metrics we used in this paper. For each metric we outline the general idea, where it came from, and how we adapted it so it could be applied to a Minecraft map. For clarity we \emph{emphasize} the actual term used for the metric throughout the paper where we define it.

\subsection{Metrics selection}

In order to select our metrics, we first went through a number of publications about automated evaluation of PCG \cite{smith2010analyzing, canossa2015towards,dahlskog2014procedural, snodgrass2017procedural, togelius2010multiobjective, summerville2018expanding, togelius2010towards, marino2015empirical, summerville2017understanding, summerville2018procedural, smith2017we, mahlmann2013modelling, pantaleev2012search, font2013card}, with a focus towards those metrics that had existing implementations. The majority of them were focused on platformers, such as Super Mario Brothers, which were useful for us, at there was a high conceptual similarity between a tile-based level and a block-based Minecraft map. The following metrics were frequently used in the platformer context : Linearity, Leniency \cite{smith2010analyzing} and Density \cite{shaker2012evolving}. Many papers also focused on patterns and how recurring certain elements are used in an artifact. There are several concepts similar to Salience, as defined by Canossa and Smith\cite{canossa2015towards}, that seemed relevant. They all capture how frequently a level object is used, how predictable its occurrences are, and how it stands out from the rest of the level.
We also looked at publications aimed at other game genres. Among the metrics we found useful in that process were particularly those focused on level design, as it is the closest in content to our artifact type. While we would have appreciated using metrics developed for others purposes, we could not find a way to apply them to our artifacts. A good example are metrics that are developed for narrative content, as they are designed to evaluate content that is intrinsically narrative (such as text or a narrative graph) \cite{purdy2018predicting, szilas2014objective} whereas the GDMC relies mostly on environmental storytelling. Before running these computational metrics, we would first need to be able to convert the 3d block-based settlement environment into pure narrative content, which falls out of the scope of this paper.
We removed metrics that were not applicable to Minecraft, such as time to finish a level, and other gameplay metrics.
Among the remaining metrics, some of them are just suggestions without any formal definition or implementation. Therefore, we also excluded them. Finally, we removed metrics that would not work with our block based representation, such as sound related metrics, or color extraction based on texture and the like. 

\subsection{Metric Development}

In the following section we describe how we adapted or developed specific metrics for this paper, often starting with the original inspiration, and then discussing the decisions we needed to adapt that metric to be applicable to a Minecraft map.

\subsection{Frequency based Metrics}

One of the earliest computational metrics for PCG in platformers was leniency~\cite{smith2010analyzing}, a measure of how forgiving a level is. The original implementation in Super Mario assigned certain elements, such as enemies, bullets, etc., a lenience score, and then counted those elements, summing up the score for the level. For platformers there are other metrics, likely inspired by leniency, that simply count enemies \cite{marino2015empirical,summerville2017understanding}. This is not that easy in Minecraft, as the enemies are entities and as such are not directly represented in the block representation and are also free to roam around and change location. I.e. you could not ensure that they meet the player at the same place, or are even still there when the player gets there. In platformers there are also other metrics that count the frequency of every tile or block \cite{marino2015empirical,summerville2017understanding}, but given the very long list of blocks in Minecraft we decided against this. However, given the high success of predicting human evaluations with frequency based metrics, such as lenience and enemy count \cite{marino2015empirical}, we wanted to include some form of frequency metric.

The specific list of tiles used in Super Mario (or similar games) is obviously domain specific, and needs to be translated to Minecraft. Here we ran into a problem as it was difficulty to track enemies, and other blocks that are dangerous (such as lava) are usually not found, nor generated in a settlement. 
On the other hand, we reasoned that certain blocks might make the game easier for the player, by providing light, or aiding the defense, or helping the player produce food to fend of starvation. So we decided to adapt the frequency based counting method to our Minecraft maps by defining, based on our domain knowledge of Minecraft, several sets of blocks which gave scores to several metrics, namely: \emph{Light, Defense, Functional Metric, Aesthetic Metric and Food}. See the appendix for list of each of the blocks contained in those metrics. We evaluate all these metrics in a similar way to leniency. For each metric, we counted the blocks that match the category and then normalized the result by dividing by the sum of added blocks. We ended up with 8 metrics. Some of them, such as the Aesthetic Metric, were a departure from the leniency idea, as they did not aim to capture a group of functional blocks that might make the game easier for the player, but to capture a different aspect of the game. In general, we wanted to see how well frequency based metrics could predict the quality assessment of humans in the different categories. 

\subsubsection{Light}
Light in Minecraft plays a major role in defense, as it prevents mobs from spawning. This specific point is one example to illustrate the functionality criteria to the GDMC judges in the evaluation guide. In addition, lighting of a settlement can be used for its aesthetics and visual narrative. MCEdit \cite{mcedit} does not let us simulate the behavior of the light in Minecraft game's engine directly, so we cannot measure how well lit an area is. Instead, the \emph{Light} metric counts every block that emits light. Light was also an aspect often raised by the human judges in the evaluation notes for the first year.

\begin{figure*}[h]
  \centering
  \includegraphics[width=0.49\linewidth]{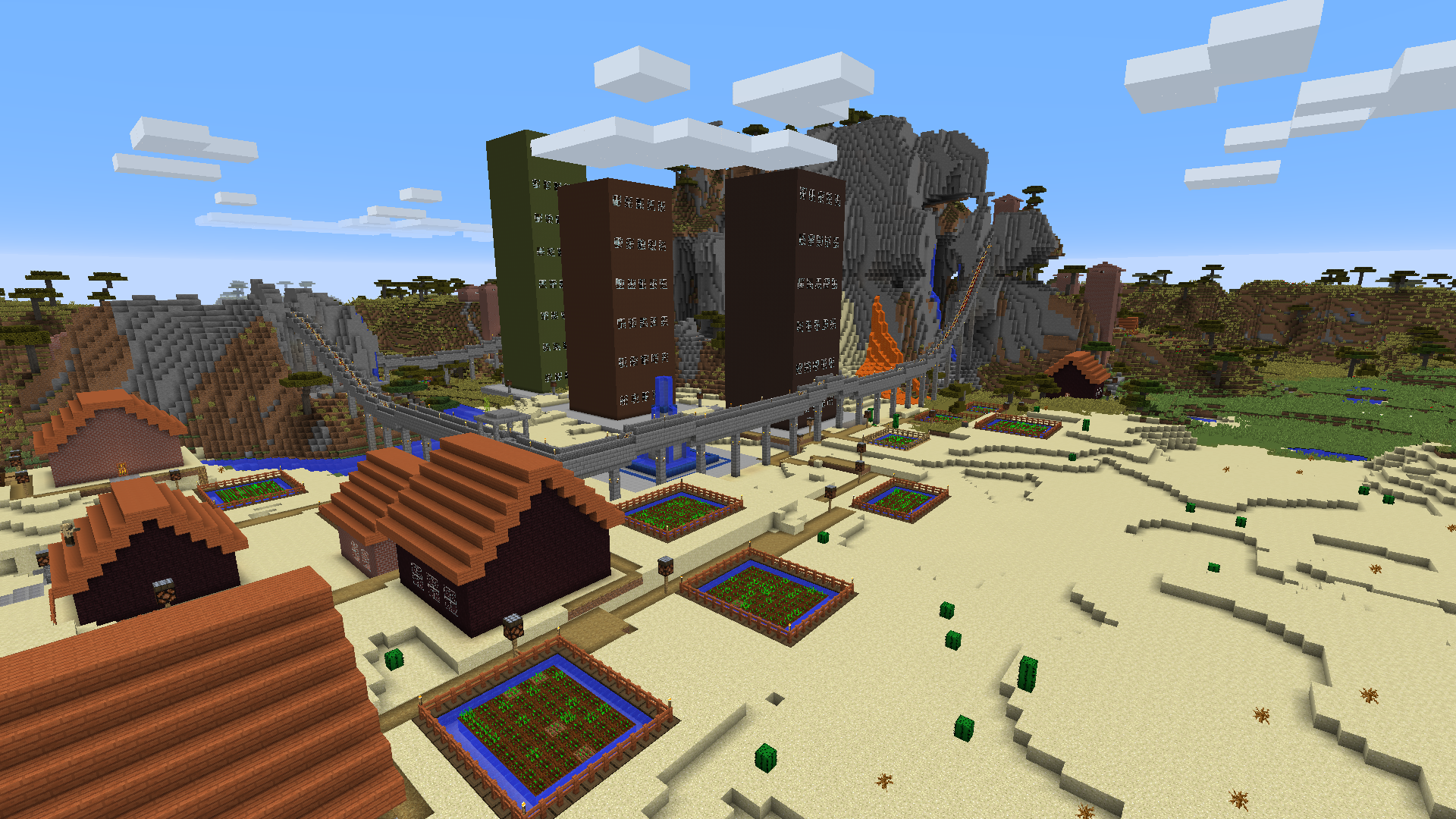}
  \includegraphics[width=0.49\linewidth]{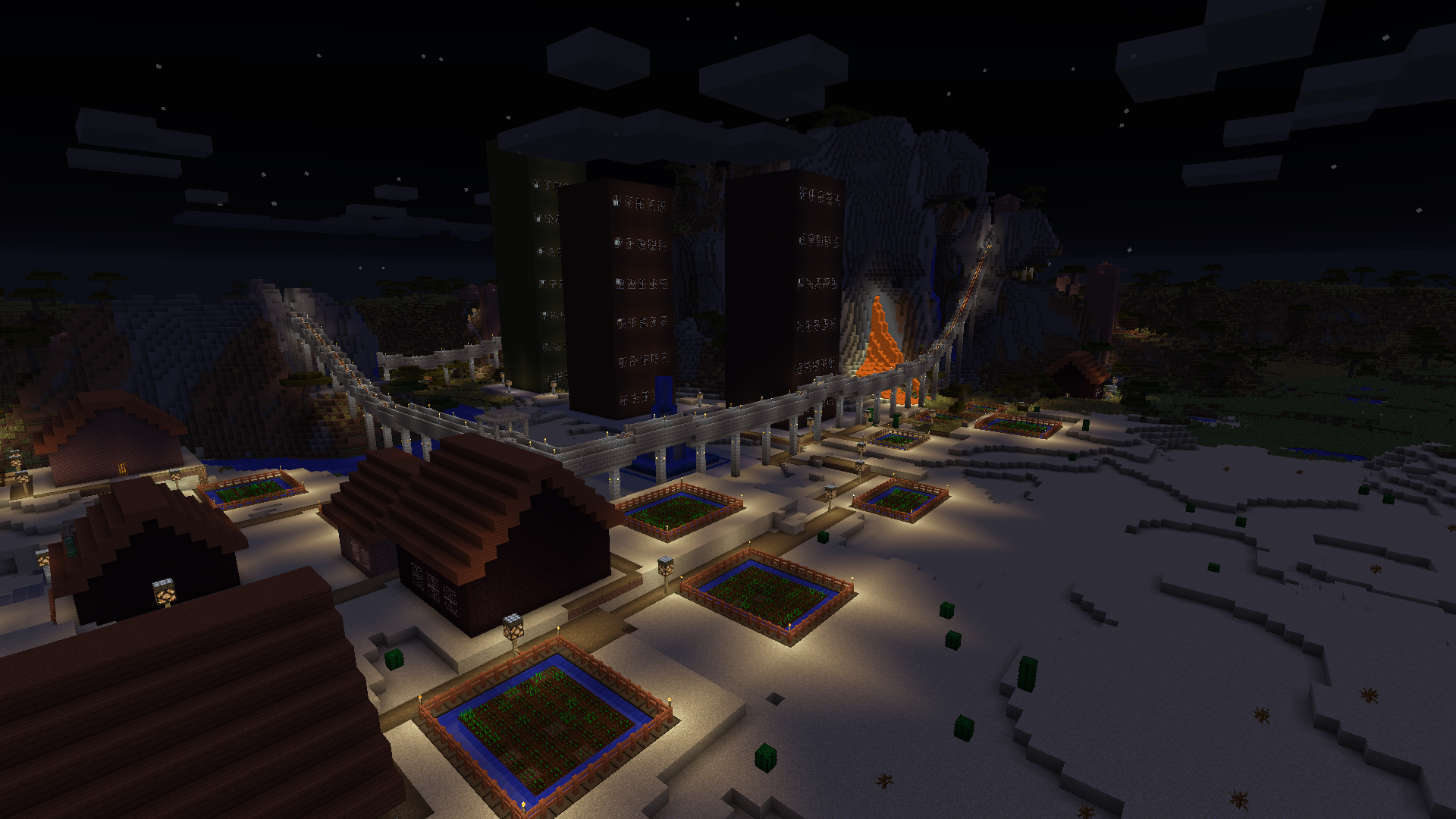}
  \caption{Example of the same scene at both day (left) and night (right) time. Scene includes several light emitting block, namely torches, lava and glowstone. Note how the light at night changes the mood of the settlement.}
 \label{ref:light}
\end{figure*}

\subsubsection{Defense}
We also count the number of blocks that could be used for defensive purposes, either in terms of narrative or in terms of gameplay. The \emph{Defense} metric includes blocks like doors, fences, etc., that can't be crossed. Also in-game traps, such as trip-wire or trapped chest. We excluded blocks that could be technically used in that regards, like stone walls, since it is not their primary function and our evaluation can't determined how they are used. We however included water and lava blocks, as their game mechanics make them useful in any defense scenario. They are the two only blocks that can inflict damage, by burning or drowning, and restrict movements.

\subsubsection{Functional Metric}
Some blocks in Minecraft serve a clear functional purpose, like beds, anvils, or chests. These are blocks that can be interacted with in the world by right clicking and then offer a specific function. One GDMC evaluation criteria, Functionality, evaluates what affordances and functionality the settlement provides. Some of the more abstract notions, such as navigability are hard to quantify, but we can count all blocks that provide interactive functionality. Note that this metric just counts every instance, so it does not check for a variety of functionality, i.e. how well a settlement covers the various different functions. However, it does indicate if the number of functional block match the size of the settlement, via the normalization. Given that this metric aims to capture the Functionality criteria we named it \emph{Functional Metric}, adding \emph{Metric} to distinguish it from the GDMC evaluation criteria we compare it with.

\subsubsection{Aesthetic Metric}
Some Minecraft blocks serve no particular purpose, but do have a special look, like glass blocks or colored wool blocks. Other blocks that have a diegetic function, but are not that useful in terms of game mechanics (e.g : Hay Bale), fall in that category. The \emph{Aesthetic Metric} captures how many of those block there are. The category contains both artificial blocks that are hard to produce, yet provide no additional function besides their looks, and natural occurring blocks that don't have function purposes (like food), but are often used to decorate settlements.

\subsubsection{Food}
Settlements are supposed to provide foods to their inhabitants. Food can be harvested from animals or plants. Our framework does not allow us to gather data on animals. Therefore, we gather the amount of edible plant blocks available in the settlement. For the \emph{Food} metric we count all blocks that are edible or involved in making edible food. 

\subsubsection{Block Type Count}
Variety is a concern addressed by several publications and numerous metrics and methods are defined in order to capture it \cite{smith2010analyzing, canossa2015towards, summerville2017understanding}. We decided to base our own variety measurement on the number of different types of elements a settlement uses, since this is a very straightforward metrics for grid-based game, such as Minecraft. This method has been used both for gameplay and aesthetic evaluation \cite{summerville2017understanding}. In platformers this computation is done simply by going through the whole grid level and for each non empty cell checking the type of object present and counting the number of unique objects. 
Minecraft has a significant amount of different blocks. For our study we counted 252 of them, based on the version of Minecraft the competition used (1.12.2). As a metric, we count how many different block types are present in a given settlement. A greater value means that the settlement uses a larger variety of blocks, which means a wider diversity of materials and objects. It could be linked to the perceived creativity or competence of the generator, as it demonstrate the ability to use, i.e. place, a larger variety of blocks, and also reflect a diversity of offered functionalities, but having a larger range of different functional blocks. 

\subsubsection{Relation to Environment}
Following on the idea of \emph{Block Type Count} and variety, we defined another metric that works similarly, but tuned for Minecraft.
Minecraft revolves a lot around the fact that you can combine blocks to create new ones. Different crafted blocks required different input blocks, forming a complex web of crafting relationships \cite{guss2021minerl} that a human player would take hours to fully explore. We compute the relation to environment as a score that we computed the following way. For each block type present in the settlement, we check if the current level provides the resources required for crafting such a block. If not, the score is not increased. If it can be crafted, the score is increased by the number of blocks that are required to be crafted, from raw materials to the actual block. (e.g. Chest blocks require Plank blocks that are crafted from Wood blocks, so the presence of Chests would increase the score by 2). This intermediate step allows us to capture how advanced a settlement is and if it has a large control of its environment, rather than using only raw materials. We voluntary exclude material obtained through mining from this evaluation, as they are quite common while not being visible from a settlement perspective. A judge or player visiting a settlement can't clearly determine if the underground of a settlement contains metal or not, but a visitor with little experience in Minecraft could know that all these materials are easily obtainable. This measure should reflect the complexity of the material used - if they were not simply placed by an editor. It should be high for those settlements that use difficult to produce crafting material but also showcase the intermediate steps to make them. It should also be high for those settlements that use intermediate materials that can be made directly from the resources present. This metric was defined to capture another guideline given to the GDMC judges in the evaluation guide, namely to see how well a settlement is adapted to the material provided by the environment. 

\subsection{Configuration based Metrics}

\subsubsection{Density}
\emph{Density} in platformers measures how many platforms are stacked on top of each other - or on average, how many platforms there are in a level of fixed length \cite{shaker2012evolving}. To adapt this to Minecraft we first have to define what we would consider a platform, i.e. a position to stand on : lets call this a \emph{surface block}. We define this based on our domain knowledge of Minecraft's game mechanics. Basically any solid block that has two empty block spaces above it is a surface block. To compute the density we count all surface blocks. This includes blocks added by the generator and the surface blocks already existing in the environment as well. This result is then normalized by the level size that we use for our experiment, i.e. the number of possible blocks in the whole environment. To get a high value here, a generator needs to build multi-story houses with several low ceiling floors, which fits nicely with the term density as used in urbanism. This value should be higher for generators that create structure with several surface blocks stacked above each other, such as the many floors of the highrises or skyscrapers seen in Fig.~\ref{ref:light}.
This is closely related to other metrics\cite{canossa2015towards} that capture how many of those surface blocks are actually reachable. But we did not implement a ``reachability'' metric due to reachability calculations involving simulating gameplay. 

\subsubsection{Filling Ratio}
The \emph{Density} metric, as computed in the platformer genre, can also be interpreted as a measure of how filled the level is with platform tiles. We made another interpretation of this metric trying to capture this aspect in Minecraft. 
Filling ratio is a simple computation of how much of the allocated space is actually used, i.e. how many of the blocks are not empty. For a similar environment, a generator that creates fewer or smaller structures will end up with a smaller filling ratio than a generator that would generate as many blocks as possible. This is measured on the overall level, so the generator might not affect the overall percentage too much, but a difference should be visible if different generators are applied to the same starting map. The one large outlier here was the generator by David Mason, who created a massive quarry inside each settlement, removing a large amount of blocks to generate a large feature. 

\subsubsection{Linearity}
Linearity is another early metric for PCG in platformers, and basically measures how well the height of the platforms in a level fits to a straight line \cite{smith2010analyzing}. As described by Smith and Whitehead, Linearity is computed by applying a linear regression to all platform's height. All platforms' height are then compared to the line, and the differences are summed up. Therefore, a low value means that most platforms are aligned and the level has a strong linearity. As opposed to the previous, count based metric it is sensitive to how things are arranged spatially, but we can re-use our definition of \emph{surface block} from the density metric, producing a list of 3d coordinates of surface blocks. To treat this as a 2d linearity problem, we projected all the surface blocks onto the two vertical planes, X and Z, matching Minecraft's ground axes. We compute both linearities based on these 2d projections, calculating the average squared distance from the regression line of best fit (least square error).

\subsubsection{Platform Size}
While our first implementation is very close to the definition of Linearity \cite{smith2010analyzing}, there is the concern that the overall linearity of the Minecraft maps is dominated by the input terrain. We aimed to design another metric that captures the basic idea of Linearity, but would be more suitable for the Minecraft map. The ability to compare the two resulting metrics potentially allowing us to determine which translation of Linearity makes more sense in context.
We again reuse the definition of \emph{surface block}. In Minecraft blocks that are adjacent and of the same height are walkable (without jumping), and so adjacent surface blocks of identical height form platforms. We then compute the size of those platforms as follows: We take every surface block (even those that are below other blocks, as long as you can still stand on them), and use a floodfill algorithm to compute which of those are connected to a single platform, i.e. are adjacent and all of the same height.
The sum of all platforms' sizes is then divided by the number of platforms, resulting in the average size of each platforms. This \emph{Platform Size} value should be large for those generators that flatten out large sections of the map, removing the natural elevation changes in the terrain, or create buildings with large floors. 

\subsubsection{Entropy}

There are several existing metrics that relate to how predictable, regular or surprising a level is. Different forms of saliency \cite{togelius2010towards} measure the existence of a few, rare and hard to predict patterns. Analyzing the distribution of features \cite{snodgrass2017procedural} and patterns \cite{dahlskog2012patterns} of an artifact have also been used to measure this. A lot of these metrics are related, as they all measure how easy it would be to predict more of the level if some of the level was known. As a proxy, we implemented a form of \emph{Entropy} metric that is closely related to the Markov chain approach of level generation \cite{snodgrass2017procedural}.

 For each block type, we look up all the block types that follow it in one dimension, and based on theses observation we compute the probability distributions of block types following each block type \cite{snodgrass2017procedural}. So, as an example, if we are going in the x-direction, we might look at the coordinates 1,1,1 and see that the block type there is ``dirt''. We then try to compute what the distribution of blocks for the coordinate 2,1,1 is based on that. This creates a conditional probability distribution where the two variables are the types of two adjacent blocks. We then compute the average entropy for all those conditional probabilities. Because the majority of blocks are actually empty blocks, we ignore them in order to avoid numerical instability. In contrast to platformer level generation, we have 3 instead of 2 dimensions. Based on the probability distributions obtained from the blocks in one particular map, i.e. settlement, we compute the entropy for each of the three cardinal directions. We initially measured them separately, but since they have a correlation greater than 0.99, we decided to average them and report them as one value. We do measure two types of entropy - \emph{Level Entropy} with the whole settlement and its environment, and \emph{Settlement Entropy}, with only the blocks added or modified by the generator. 
 
 We assume that entropy should be lower for settlements that are made of large number of repeated, templated buildings, and those that restructure the environment by paving the ground with blocks. 

\begin{figure*}[h]
\centering
\includegraphics[scale=0.35]{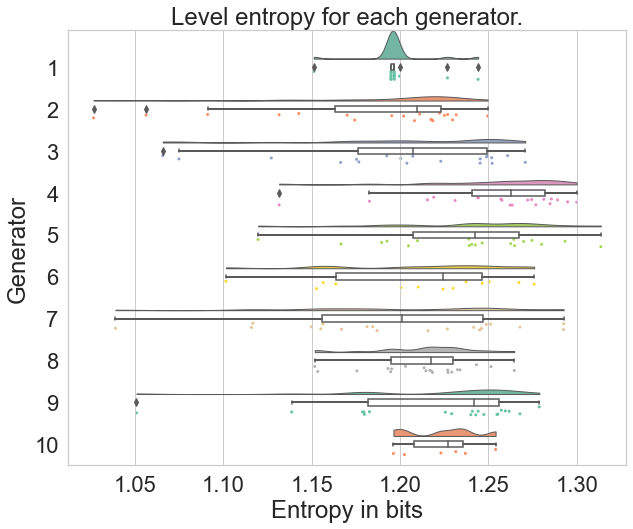}
\includegraphics[scale=0.35]{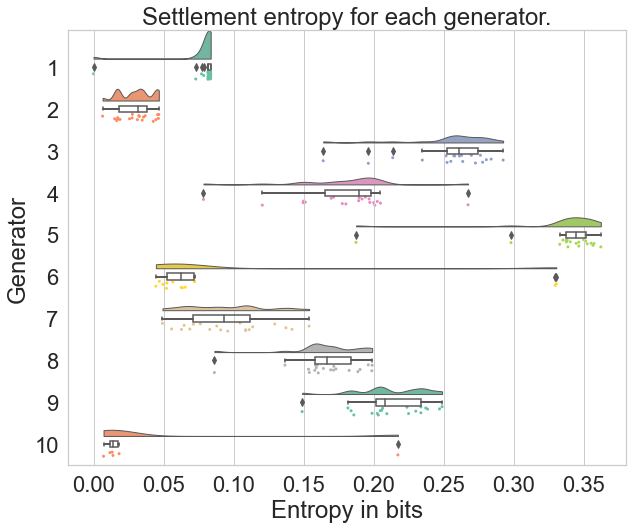}
\includegraphics[scale=0.35]{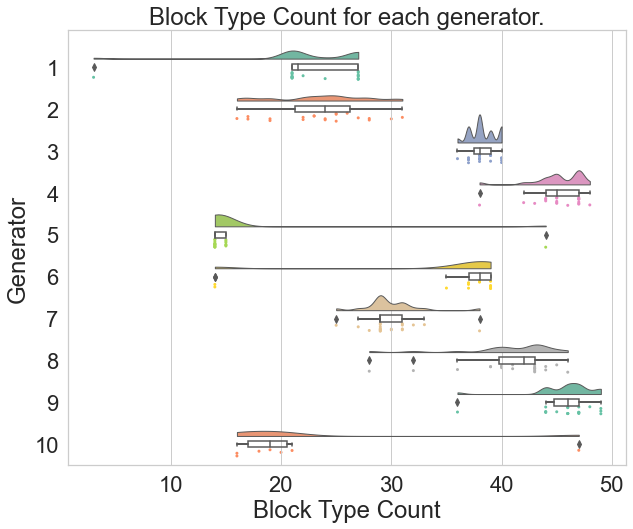}
\includegraphics[scale=0.35]{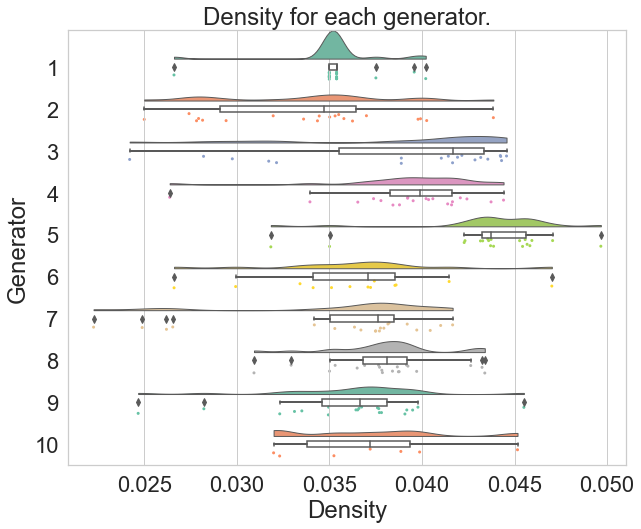}
\caption{Raincloud plot \cite{allen2019raincloud} of the distribution per generator, for the metrics \emph{Level Entropy, Settlement Entropy, Block Type Count and Density}. Plots are based on the ca. 20 samples per generator, which are displayed as drops, clouds are estimated probability density models, and box plots show the quartile ranges, with outliers being marked with circles around the dots. Comparing Block Type Count with Density we see how the high information gain metric has a good separation with lower comparative variance, while Density, the metric with the lowest information gain, has considerable overlap between all generators. Level vs. Settlement Entropy also shows how Settlement Entropy has much more distinct distributions.}
\label{fig:rainplot}
\end{figure*}

\begin{figure*}[h]
  \centering
  \includegraphics[width=0.49\linewidth]{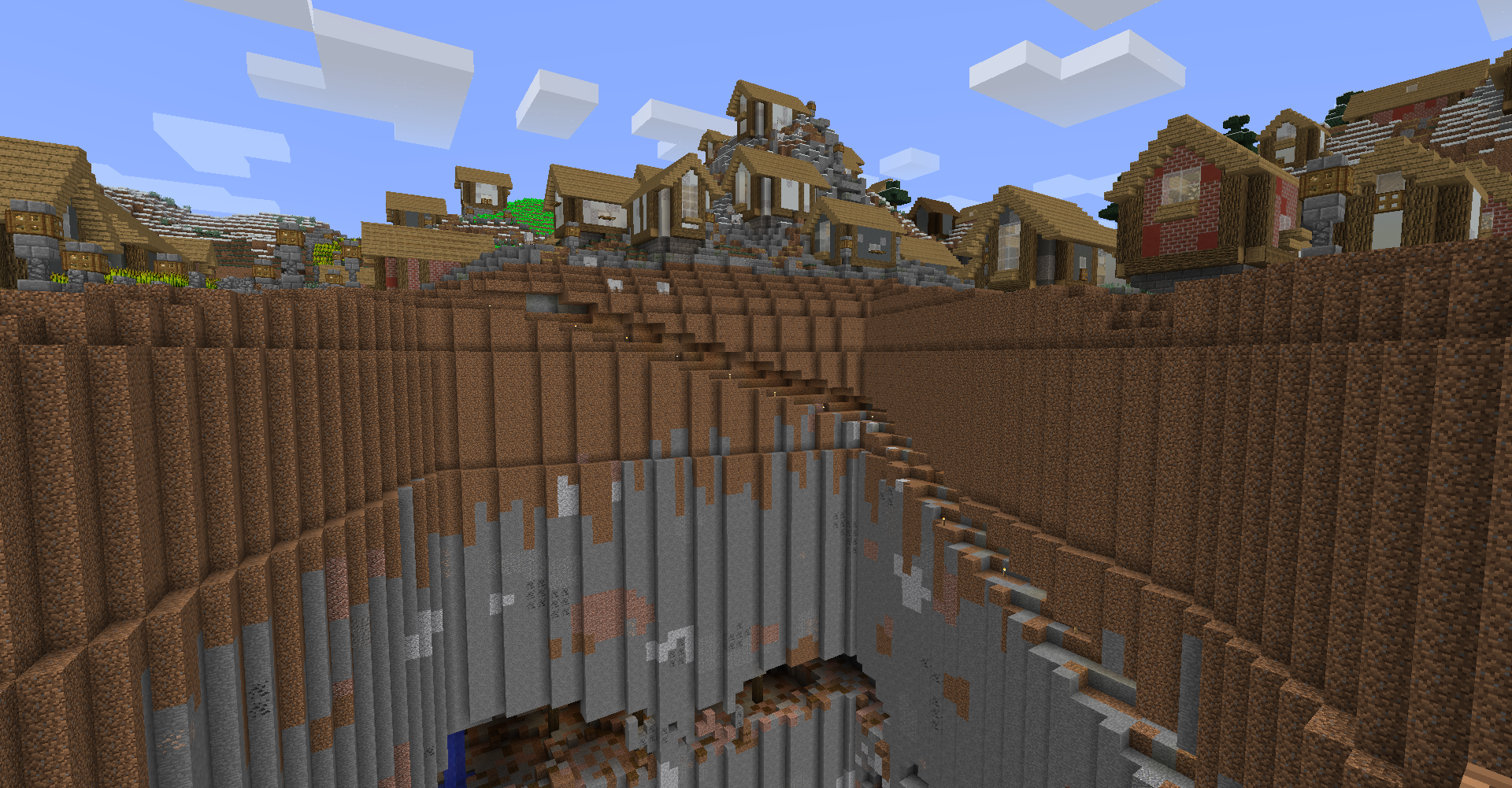}
  \includegraphics[width=0.46\linewidth]{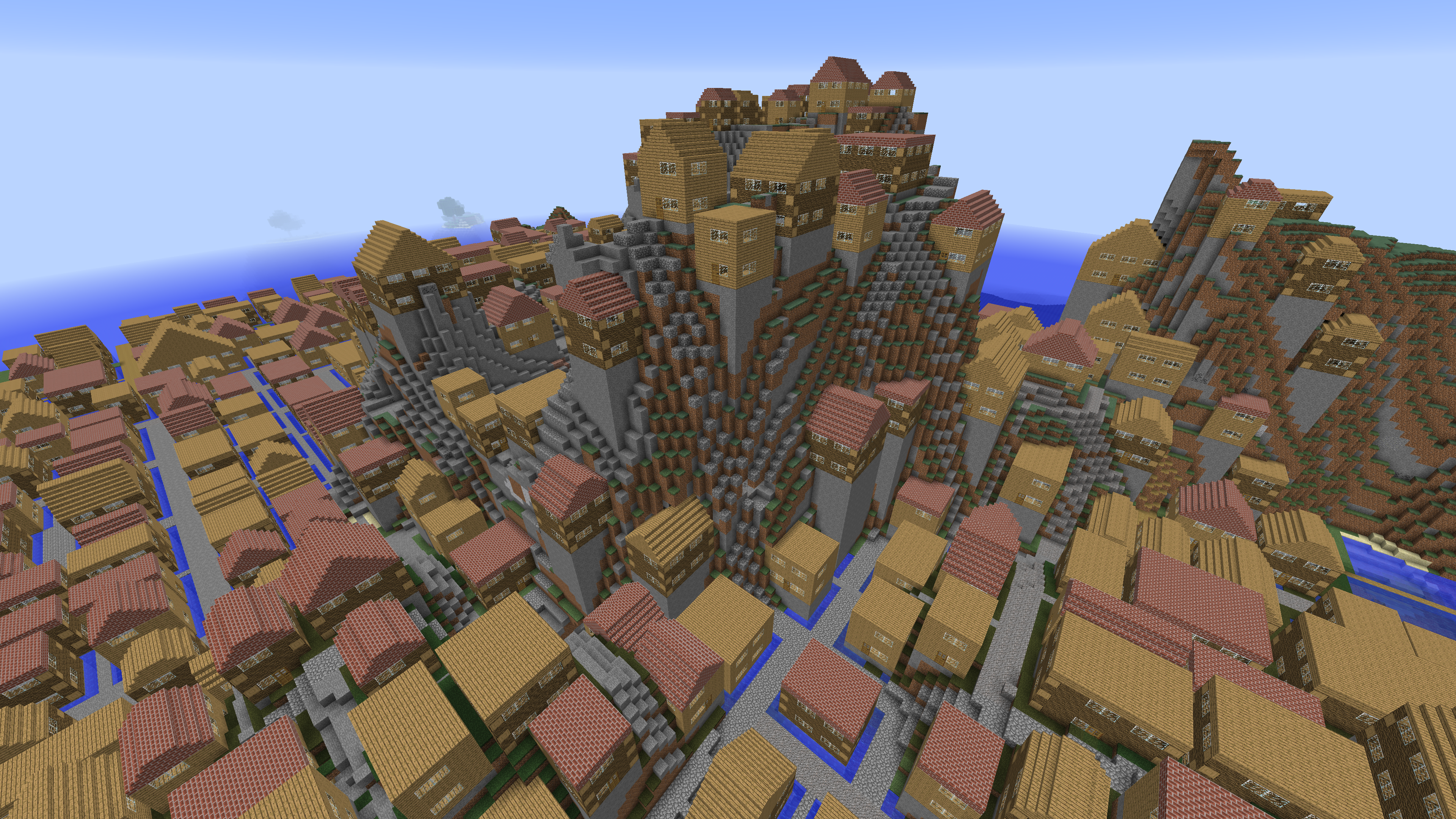}
  \caption{Example (left) of a large quarry generated by David Mason's generator, leading to this being a low outlier for \emph{Filling Ratio}. In contrast (right), the generator by Rafael Fitsch has the highest average \emph{Filling Ratio}, due to the many added houses, and also the highest \emph{Density}. It also has a very high \emph{Platform Size}, due to the many flat roofs, and massive restructuring of nearly the complete landscape into flat houses.}
 \label{ref:filling}
\end{figure*}

\section{Data collection}
We gathered our data by using entries from the GDMC through the years 2018 to 2020. To date there are 18 submissions, of which 2 have a null score and 6 are implemented in a way that doesn't match our experimental framework, i.e. they were produced with methods other than the MCEdit \cite{mcedit} based framework. That left us with 10 settlement generators that we can analyze, see the appendix for a complete list.

Each entry is a piece of code the can generate a Minecraft settlement. The entries we used were all written to work with the MCEdit framework \cite{mcedit}, which is a freely available map editor for Minecraft. It has a functionality called a filter, which allows the user to apply a piece of python code to a selected area of the map. The filter in turn needs to have a given function that gets the map and a 3d bounding box in input parameters. The filter can then read and write block information for any 3d coordinate, among other things. We produce an additional framework that allowed us to automatically apply the different filters to specific areas. We also implemented the metrics discussed, so they could be automatically applied as well. We did log which blocks where changed, as some metrics were only applied to the changed area, not the overall map - which makes it not possible to apply those metrics if only a final, resulting Minecraft map was given. 

All our simulations are achieved on the same Minecraft map that we generated beforehand with the standard Minecraft terrain generator. For each entry, we generate 20 settlements at the same, 20 random locations. These locations are picked using a hard coded seed, so every generators ran on the exact same 20 terrain locations. Two of the generators we used were crashing after a certain number of runs (respectively 12 and 6), despite several attempt to fix them. We used the averages derived from this limited data nonetheless.

After each generation, we ran our metric evaluations on the newly created settlement, and then reverted it. For each generator we then computed the average evaluations of all the 20 settlements. This gives us one sample point per generator, with a total data set of 10 samples, with metrics averaged over 20 settlements and human evaluation scores averaged between the number of judges for that year. The full set of gathered metric results is available as supplementary material\footnote{\url{https://docs.google.com/spreadsheets/d/1JNuqOfTDtDf-wnNoB-vFRsbcC7mO19e4ECK-LZ4Z4dE}}. 

\section{Analysis}

\subsection{Preliminary Data Analysis}

As a first analysis step we computed the mean, median, variance and standard deviation for the metrics for the 20 different maps, for each generator. The complete data is provide in the supplementary material, so for space reasons, we refer the reader to the data or the appendix D. We noted that our metrics operate on very different scales. While this is not a problem in general for correlation-based analyses, we were also unsure about the type of distributions, and thus chose a rank based, parameter free correlation for later experiments. 

Another concern was the question of how characteristic a given metric is for a generator. This is somewhat different from reliability - our metrics do produce the exact same results for the same artifact, but on a per generator basis there is variation based on both the underlying input map, and the inherent randomness in some of the generators. This is both evident from the data, and illustrated in Fig.~\ref{fig:rainplot}, were the values for some metrics are more varied than for others. You can also note that in some cases the variance is dependent on the generator used. To produce a quantitative and comparable measure for the quality of the different metrics, we performed an information gain analysis based on \cite{stephenson2020continuous}. This produces a comparable information value in bits, which should be high when the variance of the measured values is lower for each generator separately, then for all values measured for the metric overall. This means that the measured values do not overlap, and it would be easier to determine which generator was used, when only the measurement was provided. See the overall ranking of metrics in table \ref{Table:InfoGain}, with \emph{Relation to Environment} and \emph{Block Type Count} providing the most information. Again, Fig.~\ref{fig:rainplot} illustrates how the values for each of the different generators are not that varied, but differ between generators. Having a high value here is not necessary for a good metric, but make it more plausible that a per generator human evaluation corresponds with an average metric value for a generator. Having a lower value here, as for example the various entropies, leads to a high overlap of the measured entropyies for each specific map, and thus is much more prone to spurious correlations with low samples.  Also, for illustration, compare the two different ways we computed entropy in Fig.~\ref{fig:rainplot}, and you can see how the added entropy measure leads to more characteristic measurements that are more informative for later analysis, as the values are less determined by the input map. This kind of analysis can be helpful to design a more characteristic measure.  

\begin{center}
\begin{table*}[t]
\begin{tabular}{|c|c|}
\hline
Metric & Information Gain\\
\hline\hline
Relation to Environment	& 3.321928094887362\\
Block Type Count	& 3.129697182165249\\
Platform Size	& 0.8798718287780574\\
Food	& 0.8540055962827613\\
Light	& 0.7085994984893005\\
Functionnal Metric	& 0.6488594953924061\\
Defense	& 0.45844390596386564\\
Aestetic Metric	& 0.3650168877385864\\
Settlement Entropy Z	& 0.23900263186205217\\
Settlement Entropy X	& 0.23858966363738654\\
Settlement Entropy Y	& 0.23770177029643103\\
Level Entropy Z	& 0.13869191042675855\\
Level Entropy Y	& 0.13699304772516063\\
Level Entropy X	& 0.13675498909168393\\
Linearity Z	& 0.12417740652220077\\
Linearity X	& 0.10959274428829113\\
Filling Ratio	& 0.053946526755792146\\
Density	& 0.04437302495109918\\
\hline
\end{tabular}
\caption{Table showing the information gain for each metric for the studied set of generators. High values indicate metrics were the variance of the samples of one generator is smaller than the overall variance, i.e. the metric is characteristic for a generator and gives \emph{Information Gain} of bits information about which generator produces the sample.}
\label{Table:InfoGain}
\end{table*}
\end{center}

\subsection{Correlation Human Scores vs. Metrics}

Next we wanted to know if the metrics have a relationship to the human evaluation scores. To test for this, we computed a range of Spearman's rank correlation looking at the possible relation between the mean average of all metric evaluations vs. the mean averages of the various human evaluation scores from GDMC for each generator. We chose a non-parametric rank based correlation as we could not guarantee that each metric is normally distributed. In order to clarify the somewhat low statistical power of our sample size, we also computed the p-value of each correlation. 

We could successfully reject the null hypothesis of no linear dependency for only 5 correlations, based on a cut-off at $p=0.01$ : between Functional Metric and Jury score ($p=0.002$), Functional Metric and Functionality Score ($p=0.002$), Food and Jury Score ($p=0.004$), Food and Functionality Score ($p=0.005$), and Block Type Count and Narrative ($p=0.005$).

\subsection{Discussion}

\begin{center}
\begin{table*}[t]
\begin{tabular}{|c|S| S| S| S| S|}
\hline
 Metrics & Jury Score & Adaptability & Functionality & Narrative & Aesthetic\\[0.5ex] 
\hline\hline
Level Entropy & -0.26 & -0.26 & -0.37 & 0.02 & -0.16 \\ 
\hline
Settlement Entropy & 0.22 & -0.14 & 0.32 & 0.2 & 0.32\\
\hline
Light & 0.77 & 0.49 & 0.63 & 0.72 & 0.73\\
\hline
Defense & 0.42 & 0.07 & 0.55 & 0.28 & 0.47\\
\hline
Functional metric & 0.84 &  0.75 &  0.85 &  0.43 & 0.65\\
\hline
Aesthetic metric & 0.31 &  0.10 & 0.39 & 0.12 & 0.32\\
\hline
Food & 0.82 & 0.54 & 0.81 & 0.58 & 0.71\\
\hline
Relation to Environment & 0.73 & 0.42 & 0.65 & 0.61 & 0.61\\
\hline
Block Type Count & 0.67 & 0.28 & 0.55 & 0.81 & 0.70\\
\hline
Density & -0.24 & -0.33 & -0.33 & -0.04 & -0.18\\
\hline
Platform Size & -0.33 & -0.18 & -0.41 & -0.27 & -0.29\\
\hline
Linearity X & -0.19 & -0.14 & -0.03 & -0.37 & -0.24 \\
\hline
Linearity Z & 0.15 & -0.27 & 0.41 & 0.24  & 0.30\\
\hline
Filling Ratio & -0.14 & -0.03 &  0.03 & -0.13 & -0.14\\ 
\hline
\end{tabular}
\caption{Spearman’s Rank Correlations between metrics and GDMC scoring. Bold correlation have a p-value below $0.01$. Due to fixed sample size p-values directly depend on correlation value.}
\label{Table:1}
\end{table*}
\end{center}

Our results point out that overall, most of the metrics have a low correlation with the Jury's scores, below $0.5$ (See Table 2).

Functional metric and Food have a strong correlation with the Functionality score, respectively $0.85$ and $0.81$. That can be easily explained as providing services is one of the main criteria in this category. Those metrics are doing a good job at capturing this, however they are deeply connected with Minecraft and its game mechanics. Those three metrics have also a correlation of $0.84$ and $0.82$ with the Overall Score, which led us to think that they are playing a key part in the evaluation of a good settlement. But this could also be due to the fact that Functionality plays a major role in the overall score, as shown in Table 4.
Other metrics that are linked to Minecraft's mechanics do not have the same impact. However, our computations of Light and Defense are still limited. More robust implementations of these could actually lead to a good method for capturing how functional a settlement is.  

Block Type Count is the metrics that, based on our observation, performs the best for Narrative. It has the highest correlation with this category ($0.81$) with a low p-value. Block Type Count could actually be a good metric of how a generator use the different possibilities provided by Minecraft in order to create the most evocative artifacts. The relation between these two elements has yet to be investigated, but this is actually an interesting track in the development of environmental narration metrics, which we are still lacking.

We would also like to point out that while most of the scoring categories have some sort of correlation with our metrics, none of them successfully correlate with Adaptability and Aesthetic. Adaptability and Aesthetic are two high concepts that seems hard to be approximated using quantitative metrics.

Looking critically at the p-values we see that only 5 of them are below 0.01, and we should point out that since we used a very exploratory approach and checked about 70 possible correlations, that this is a relatively weak result, and there is a possibility for chance results. This is due to the low number of samples (10) - with the bottleneck for this approach being the number of generators that exist. Given promising candidate measure it would be possible to produce a statistically more powerful dataset. For example, existing generators could be used with some diversity metric to generate several, ideally different settlements each, and then all of them could be given to humans to evaluate - and we could compute the metrics for each of those maps at time of generation. Working with existing human scores limits us to those generators we can currently run, but we could expand this number as future GDMC competitions produce more generator-score sample points. 

We should also point out that while all metric evaluation for the generators are done on the same 20 map locations, the human scores are based on having the generators applied to 3 different maps, which changed between years. While this is not ideal, we posit that the metric values are a property of the generator itself, which we try to extract by repeated measurements on random location. But we wanted to ensure the direct comparison between the generators of different years, and as such chose to go with a new set of consistent maps for all generators. 

\subsection{Correlation between Metrics}

We also performed a correlation analysis between the different metrics to see how they relate to each other (Table 3).

The correlations of Settlement Entropy with our Aesthetic metric ($0.89$) shows that this metric is related to the structures of the settlements and the patterns within them. The occurrence of doors, windows, and other aesthetic blocks, affects the Settlement Entropy by adding a wider range of thereby harder to predict blocks.

The Block Type Count correlates with Light ($0.88$), which is not surprising as a large panel of blocks emit light. Using different light sources lead to a likely increase of the number of block types used.
For similar reasons, we also observe a strong correlation ($0.82$) between Light and Functional metrics. Several Functional Blocks emit Light, so this result was expected.

We also looked at how the scoring categories relate to each other, following up on the question if overall impressions bias the sub-scores. The fact that the overall score is strongly related to the sub-scores is unsurprising, since it is a linear combination of those. Apart from that, we can see that there is quite a bit of dependence between the scores - with Adaptability and Functionality forming one, and Narrative and Aesthetics forming another closely related pair. It is unclear if this is due to an overall bias in the judges evaluation, i.e. the judge biases their sub-category scoring on the overall impression, or if the quality of the generator is a confounding variable, i.e. those generators that are good at dealing with one aspect are also more likely to deal well with another aspect.

\begin{center}
\begin{table*}[t]
\scalebox{0.87}{
\begin{tabular}{|c| S| S| S| S| S| S| S| S| S| S| S| S| S|} 
\hline
& S.E. & Lig. & Def.e & F.M. &  A.M & Food & RtE & B.T.C & Den. & P.S. & Lin.X & Lin.Y &F.R.\\ [0.5ex] 
\hline
Level Entr. & 0.41 & 0.10 & -0.15 & -0.21 & 0.49 & -0.50 & -0.25 & 0.16 & 0.85 & 0.53 & -0.81 & -0.24 & -0.36\\ 
\hline
Settlement Entr. & & 0.43 & 0.66 & 0.24 & 0.89 & 0.16 & 0.19 & 0.32 & 0.64 & 0.09 & -0.26 & 0.45 & -0.10\\
\hline
Light & & & 0.26 & 0.82 & 0.37 & 0.62 & 0.66 & 0.88 & 0.13 & -0.41 & -0.32 & 0.18 & -0.39\\
\hline
Defense & & & & 0.36 & 0.50 & 0.62 & 0.52 & 0.19 & 0.19 & -0.36 & 0.05 & 0.67 & -0.09\\
\hline
Functional metric & & & & & 0.28 & 0.78 & 0.71 & 0.62 & -0.15 & -0.54 & -0.10 & 0.14 & -0.28\\
\hline
Aesthetic metric & & & & & & 0.16 & 0.14 & 0.28 & 0.61 & 0.38 & -0.47 & 0.27 & -0.14\\
\hline
Food & & & & & & & 0.88 & 0.58 & -0.30 & -0.62 & 0.08 & 0.45 & -0.33\\
\hline
Relation to Env. & & & & & & & & 0.70 & -0.08 & -0.49 & -0.12 & 0.21 & -0.53\\
\hline
Block Type Count & & & & & & & & & 0.07 & -0.27 & -0.49 & 0.27 & -0.44\\
\hline
Density & & & & & & & & & & 0.39 & -0.56 & -0.08 & -0.50\\
\hline
Platform Size & & & & & & & & & & & -0.44 & -0.41 & 0.07\\
\hline
Linearity X & & & & & & & & & & & & 0.13 & 0.50\\
\hline
Linearity Z & & & & & & & & & & & & & 0.32\\		
\hline
\end{tabular}
}
\caption{Spearman's correlation coefficient between metrics, top row abbreviated. Bold correlation have a p-value below $0.01$.}
\label{Table:3}
\end{table*}
\end{center}

\begin{center}
\begin{table*}[t]
\begin{tabular}{|c|R R R R|} 
\hline
& Adaptability & Functionality & Narrative & Aesthetic\\ [0.5ex] 
\hline\hline
Jury Score & 0.82 & 0.87 & 0.73 & 0.93\\ 
\hline
Adaptability &&	0.67 &	0.37 & 0.64\\ 
\hline
Functionality &&& 0.44 & 0.76\\
\hline
Narrative &&&& 0.88\\
\hline
\end{tabular}
\caption{Spearman's rank correlation between scoring categories.}
\label{Table:4}
\end{table*}
\end{center}

\section{Conclusion}

For the conclusion we can return to the three questions that drove this research in the first place.

How well do existing PCG metrics generalize to other domains, specifically to the domain of Minecraft settlements? This question decomposes into two parts - can those metrics be applied and are their results meaningful? The first part turned out to be relatively unproblematic - as our report on how we adapted the metrics outlined. We did have to introduce some domain knowledge, as most count based metrics, with the exception of Block Type Count, required a domain based labeling of blocks, indicating if they are dangerous, functional, etc. In general, this seems to be a requirement in order to make them fit for any game. Furthermore, we should also point out that the jump from a 2d tile based domain to a 3d block based domain is not that large. The metrics that relied on this, such as entropy and filling ration, were easy to adapt, and the modifications seemed principled. Density and Linearity required some modifications - as they could have been applied in a more literal sense, but this felt problematic, as they did implicitly encode some game mechanics, such as traversal through a level, so we ended up changing them to better reflect what we think linearity and density should mean, i.e. introduce the idea of a surface block a player could stand on. Even with those adaptations, we should note that while those metrics can be applied, already the information gain analysis indicates that some of them, such as Linearity or Density, might not be particularly meaningful as their results are very noisy. 

Finally, it should also be pointed out that a whole lot of potential metrics were left out. Literature was quite full with suggestions of what should be measured, but for many items, these were never implemented, as they were probably already difficult to formalize for the domain they were intended for. Others were left out because they did not match the artifact type we evaluated - such a music evaluation. A Minecraft settlement seemed to mostly correspond to a level, but this also raises the question of how we would determine what types of artifacts are comparable, or what components are causally important if we evaluate whole games?

If those metrics remain meaningful in terms of human experience brings us to the next introductory question, are those metrics grounded in human experience. The results in this regard are somewhat lackluster, in part due to an unfortunately small sample size. We do see some moderate and major correlations between the scores and the human evaluations, but few of them have significant p-values. Two of those, the Light and Functional relation to the Functionality score, are dependent on providing domain independent information, which speaks against their generality. Interestingly, in a paper that looked at the human grounding of metrics for Super Mario levels both the original lenience, and enemy count, where also, by far, the most indicative values for the different quality assessments \cite{marino2015empirical}. So the principle of count based metrics seems to robustly generalize between those two. Block count, the other significant metric, seems to have a positive relationship with the Narrative score, but has no direct equivalent with platform level generation. Here it would be interesting to see if this generalizes back to level generation - as a different kind of entropy. 

Finally, how robustly this approach deals with more complex content is hard to answer conclusively. We did note that while some metrics were able to indicate which of the predicted settlements were rated higher by humans, those metrics did not seem to be suitable to optimize by on its own. Having a settlement with a maximum number of light emitting cubes, or using every possible Minecraft block is not sufficient, or even a good idea. In contrast, prior work \cite{yannakakis2011experience} for Super Mario has used metrics to adapt levels to user preferences. But in Super Mario playability, as evaluated with an automatic game play agent  \cite{Baumgarten2010}, can be used as a feasibility test to ensure that the search space of levels is restricted to those that are ``okay'', and the metrics are then used for further refinement. Something similar could be done with Minecraft settlement generators. If a generator could ensure that it only generated ``okay'' settlement, then parameters of the generator could be fine-tuned to make a better settlement.  But it is, at least to us, unclear how an algorithmic evaluation that would ensure this, similar to playability for platformers, would look like. 

The fact that a Minecraft settlement is a form of holistic PCG, combining several facets \cite{Liapis2019} of PCG in the same artefact also raised some questions when we selected metrics based on type. There is a high similarity with platformer levels, and so nearly all metrics used where suited for that type of content. Just treating it as a level though ignores several of those facets. While there are no readily available metrics for things such as how evocative the narrative content of an artifact is, this also hints at a deeper problem for PCG evaluation. How do you actually automatically determine what kind of content you are dealing with? This would come up if we wanted to move further towards evaluating games as a holistic artifact. One approach here would be to focus on developing metrics that operate on a more generalizable interface, such as looking at the experience quantified by sensory perception and actions performed by the player. Ultimately, this could lead to a systematic accounting of how to link PCG artifacts with sensory perception and ultimately with human experience. While this is beyond the scope of this paper, we believe that connecting human evaluation to existing formalized evaluations of PCG artifacts is a good first step towards this goal.   

\section{Future Work}

One obvious way to expand upon this work would be to increase the statistical power of the analysis with a larger sample size - either by waiting for the GDMC to produce more MCEdit suited generators and evaluations, or by conducting additional human evaluation ourselves on a wider range of generated settlements. This could provide further evidence, and allows us to have more clarity if the moderate correlation we see here are coincidental or not. 

The other way to expand upon this work would be the development of novel and better metrics. We already discussed that there are certain aspects of the settlements that are not captured by any of the existing metrics. Furthermore, the metrics we looked at here applied directly to the artifacts, and did not consider interactions or gameplay experiences. So using an automatic game play agent \cite{guss2021minerl} and recording some simulated interaction with the game maps, similar to the interactions the judges might have, could provide better insights. In general, this could move us closer to developing more generalizable metrics that operate at the interface between the human and the game, and are capable of providing an evaluation on any type of computer game, purely by analyzing the sensor and actuator interaction with the game.  

\newpage
\bibliographystyle{ACM-Reference-Format}
\bibliography{sample-base}

\newpage
\appendix
\section{Generators Used}
List of used GDMC entries ( from \url{http://gendesignmc.engineering.nyu.edu/results})\\

\noindent 1 - Filip Skwarski - 2018 -

\url{https://drive.google.com/open?id=19T09qQhdIYWs31TLLaGaZogH-ye_P_HA}\\
2 - Rafael Fritsch - 2018 - @UrsaMinorDev -

\url{https://drive.google.com/open?id=1YpnRgMHdQlcYu-0UjENYhrfYm576rty-}\\
3 - Adrian - 2019 - @TheWorldFoundry -

\url{https://drive.google.com/open?id=19jRqunPUP6kYVKSDxK8py_iGR1rW-_Rq}\\
4 - Julos14 - 2019 -

\url{https://drive.google.com/open?id=1KpZ-Rkt2cFTejWILoQqF9P3rb9StImw7}\\
5 - Filip Skwarski - 2019 -

\url{https://drive.google.com/open?id=1LmL3P7fHdIgOWxA02-4TulKzof2qM6CN}\\
6 - David Mason - 2020 -

\url{https://drive.google.com/file/d/1lVUHBibnWSWRdyLsIgqpO_rcrB6et5hn}\\
7 - The World Foundry - 2020 - @TheWorldFoundry -

\url{https://drive.google.com/file/d/1oI16cHurokOE3Z2z7APwhwE9HyapbiJ2}\\
8 - The Charretiers - 2020 -

\url{https://drive.google.com/file/d/1AwOBafnulNNaHgdXZb4d8dX4osL3gni7}\\
9 - Troy, Ryan, and Trent - 2020 -

\url{https://drive.google.com/file/d/1FEywsza2haHKBab_tkXLyWiXir_XqcXA}\\
10 - Grupo Delu - 2020 -

\url{https://github.com/Grupo-DELU/DELU-MINECRAFT/releases/tag/v1.0}\\

\section{Instructions for Judges}
\label{ref:instructions}
The following instructions were part of the brief given to the judges of the GDMC competition:

\subsection{Score Range}

The idea is to score the entries in each of those categories, to encourage participants to think on how to tackle each of those problems. Each category can score from 0 to 10. 

\begin{itemize}
    \item 0 points means that the resulting design shows no consideration of that particular criterion at all. 
    \item 1 - 4 points means that there are some aspects in which the criterion is addressed
    \item 5 points indicates a performance, in that area, that could be from a naive human. At this point, you would not be surprised if this was built by a human. 
    \item 6 - 9 points indicates an expert level human performance, over a longer time, possibly a group effort. So, we are talking about a group of city planners and architects designing a Minecraft settlement over the course of a year. The higher end of the point scare here should mean a work that would possibly win a design prize. 
    \item 10 points - superhuman performance - this is so good, you are surprised if this could be generated by a dedicated group of expert humans.
\end{itemize}

\subsection{Scoring Categories}

In our competition description we announced that we would supply the judges with questions to illustrate each category. This section will contain short descriptions, and those guiding questions. Keep in mind that those questions are just illustrative, and not exhaustive. Each category is more complex, and there are ways to address these issues that go beyond those questions. Consequently, the questions are not a checkbox list - so if there are 5 questions, you should not just give 2 points for each. You should rather ask, how well did the AI perform here compared to a regular human. 

\subsubsection{Adaptability:}
Adaptability is about how well a given settlement fits into the map and the surrounding terrain. Is the design shaped by the given map, and is the map in turn shaped by the settlement. The challenge here is to not just generate something that can be put on any map, but to generate something that reflects the input (the map). 

\begin{itemize}

    \item Do the structures in the settlement adapt to the terrain?
    \item Do the structure in the environment reflect the environment, i.e. usage of available material, adaptation to the biome?
    \item Does the settlement take advantage of terrain features or compensate for problems with the terrain?
    \item Are the settlements different in reaction to the different initial maps?
    \item Are there any other ways in how the settlement adapts to the given maps?

\end{itemize}

\subsubsection{Functionality:}
Settlements in general, and in Minecraft in particular, are not just aesthetic artefacts but also provide functionality in different forms. There are different kinds of functionality. First, there are issues that are not even dependent on Minecraft, like how easy it is to walk through the settlement, or how easy it is to navigate. There are also functions that are more closely tied to the game Minecraft, like keeping the monsters out, or providing enough light so monsters don’t spawn, or making food and crafting stations accessible to the player. This is definitely a category where the list of questions is not exhaustive - it is possible to provide extra affordances that we have not thought of. 

\begin{itemize}
\item Does the settlement provide protection from danger?
\item Does it keep mobs from spawning?
    \item  Does it keep mobs out?
    \item  Protection from other environmental dangers?
    \item  Is the settlement accessible to a player avatar in survival mode?
    \item  Can you walk everywhere?
    \item  Does the settlement provide faster modes of transport?
    \item  How easy is it to find your way around?
    \item  Does the settlement provide the player with additional affordances?
    \item  Does the settlement make ressources easy to obtain?
    \item  Is there an easy way to get food?
    \item  Does the settlement provide functionality to the villagers?
    \item  Does the settlement reflect the embodiment of the player avatar?
    \item  Is it appropriately scaled?
\end{itemize}

\subsubsection{Evocative Narrative:}
In real life settlement tell stories about the people who build the, about their lives, their culture, and their history. Settlements are also living testaments to the history that shaped them. When we look at settlements in reality, or even those designed by humans for games, we get a sense of what this settlement is about, or how it was shaped over time. The challenge here is to automatically produce a settlement that evokes a distinct story - and ideally one that is adapted to the underlying input, the map. 

\begin{itemize}
    \item  Is the settlement evoking an interesting story?
    \item  After looking at the settlement, could you give a short description of what this settlement is about that sets it apart from other settlements?
    \item  Is it clear what the function of the settlement is?
    \item  Does this function make sense in regards to the terrain and environment it is in? I.e. is the logging camp in a forest, the harbour town at the sea?
    \item  Is the functionality of the settlement supporting this narrative function? I.e. does the fortified frontier settlement have functioning walls, is the farming village equipped with functioning fields?
    \item  Does the final settlement give any indication of how the settlement developed?
    \item  Is it possible to look at the settlement and imagine in what order things were built, or what stages the development of the settlement took?
    \item  Is there an indication of the history of the settlement evident in the structure?
    \item  Are there any convincing and consistent allusion to human cultures or specific points in history that the settlement is modelled after?
    \item  Does the settlement have a culture - either fictional or historical, that is evident from the settlement?
    \item  Do you know things about this culture just by looking at the settlement?
\end{itemize}

\subsubsection{Maintaining Aesthetics:}
The last criterion is not necessarily about beauty, but about a consistent look. The challenge here is not only to produce something that looks like good design, but to do so, while also addressing the other challenges. It is somewhat simple to design a really great looking house, and just copy it down several times. But having an algorithm design houses and have them look well designed is a different question. This category contains a lot of elements that humans might get right without thinking about it, like building houses that are visually balanced, or well proportioned. In its corollary, this category is also about avoiding those jarring, strange artefacts that procedural design generates that would never be made by a human. 

\begin{itemize}
    \item  Does the settlement look good?
    \item  Is there a consistent look to the settlement? Does it appear that all structures belong to the same settlement?
    \item  Is there an appropriate level of variation in the existing structures?
    \item  Are there any jarring features that make the settlement look unbelievable?
\end{itemize}

\section{Block list}
\subsubsection{Functional Blocks}

Bed, Bookshelf, Torch, Chest, Crafting Table, Furnace, Enchantment Table, Profession Stand, Cauldron, Beacon, Anvil.

\subsubsection{Aesthetics Blocks}

Glass, Tall Grass, Dead Shrub, Dandelion, Popppy, Torch, Sign (Block), Sign (Wall Block), Jack-O-Lantern, Stained Glass, Iron Bars, Glass Pane, Mycelium, Lily Pad, Head Block, Stained Glass Pane, Hay Bale, Carpet, Sunflower

\subsubsection{Food Blocks}

Leaves, Brown Mushroom, Red Mushroom, Wheat, Sugar Can, Cake, Melon, Pumpkin Vine, Melon Vine, Vines, Cocoa Plant, Carrot, Potatoes

\subsubsection{Defense Blocks}

Water, Lava, TNT, Fire, Wood Door, Ladder, Iron Door, Fence, Trapdoor, Fence Gate, Tripwire Hook, Tripewire, Trapped Chest, Dropper

\subsubsection{Light Blocks}

Lava, Torch, Fire, Furnace (Smelting), Redstone Torch, Glowstone, Portal, Jack-O-Lantern, Beacon, Lantern, Redstone Lamp (On), Enchantment Table

\newpage
\section*{Quantitative data}

\begin{center}
\begin{table*}[b]
\scalebox{0.8}{
\begin{tabular}{|c| c| c| c| c| c| c| c| c |c| c ||c |} 
\hline
&  1&2&3&4&5&6&7&8&9&10 & Global \\
\hline
L.E. & $1.20$ & $1.18$ & $1.20$ & $1.25$ & $1.24$ & $1.21$ & $1.20$ & $1.21$ & $1.22$ & $1.22$ & $1.21$ \\ 
\hline
S.E.& $7.81E-2$ & $3.03E-2$ & $2.56E-1$ & $1.76E-1$ & $3.45E-1$ & $5.88E-2$ & $9.77E-2$ & $1.71E-1$ & $2.17E-1$ & $1.34E-2$ & $1.59E-1$\\
\hline
Lig. & $3.89E-5$ & $1.18E-5$ & $1.34E-4$ & $1.23E-4$ & $0$ & $2.60E-5$ & $6.50E-5$ & $7.39E-5$ & $1.86E-4$ & $2.60E-5$ & $7.36E-5$ \\
\hline
Def. & $4.08E-5$ & $1.01E-4$ & $1.68E-4$ & $4.82E-5$ & $1.47E-4$ & $6.37E-5$ & $1.57E-4$ & $1.37E-4$ & $1.05E-4$ & $8.45E-6$ & $1.06E-4$ \\
\hline
F.M.& $3.89E-5$ & $1.24E-5$ & $1.21E-4$ & $1.67E-5$ & $0$ & $2.60E-5$ & $6.50E-5$ & $7.53E-5$ & $2.20E-4$ & $2.60E-5$ & $6.44E-5$ \\
\hline
A.M. & $1.50E-4$ & $2.35E-5$ & $1.80E-4$ & $1.80E-4$ & $1.79E-3$ & $4.17E-5$ & $1.06E-4$ & $7.23E-4$ & $2.89E-4$ & $4.12E-5$ & $3.91E-4$ \\
\hline
Food & $6.76E-5$ & $1.20E-4$ & $7.72E-4$ & $6.46E-5$ & $0$ & $6.90E-5$ & $1.23E-4$ & $1.72E-4$ & $1.21E-4$ & $6.10E-7$ & $1.66E-4$ \\
\hline
RtE & $3.84E+1$ & $4.02E+1$ & $4.92E+1$ & $4.20E+1$ & $3.18E+1$ & $4.73E+1$ & $4.69E+1$ & $4.81E+1$ & $4.36E+1$ & $3.23E+1$ & $4.25E+1$ \\
\hline
B.T.C & $2.27E+1$ & $2.36E+1$ & $3.83E+1$ & $4.55E+1$ & $1.43E+1$ & $3.79E+1$ & $2.95E+1$ & $4.11E+1$ & $4.64E+1$ & $1.83E+1$ & $3.25E+1$ \\
\hline
Den. & $3.53E-2$ & $3.39E-2$ & $3.90E-2$ & $3.90E-2$ & $4.37E-2$ & $3.56E-2$ & $3.55E-2$ & $3.79E-2$ & $3.59E-2$ & $3.81E-2$ & $3.74E-2$ \\
\hline
P.S. & $8.09$ & $7.75$ & $6.78$ & $8.31$ & $1.22E+1$ & $7.90$ & $7.60$ & $8.47$ & $7.61$ & $8.03$ & $8.31$ \\
\hline
Lin.X. & $1.70E-2$ & $1.49E-2$ & $1.39E-2$ & $7.76E-3$ & $9.28E-3$ & $1.07E-2$ & $1.13E-2$ & $4.85E-3$ & $7.76E-3$ & $8.75E-3$ & $1.08E-2$ \\
\hline
Lin.Y. & $5.42E-3$ & $9.82E-3$ & $9.29E-3$ & $7.27E-3$ & $8.23E-3$ & $6.40E-3$ & $8.06E-3$ & $7.30E-3$ & $9.31E-3$ & $3.87E-3$ & $7.83E-3$ \\
\hline
F.R. & $1.73$ & $1.72$ & $1.67$ & $1.69$ & $1.72$ & $1.71$ & $1.72$ & $1.68$ & $1.72$ & $1.69$ & $1.70$ \\
\hline
\end{tabular}
}
\caption{Mean average of each generator per metrics (abbreviated). Corresponding generator as top row, see list in Appendix A for details.}
\label{Table:5}
\end{table*}
\end{center}

\begin{center}
\begin{table*}[b]
\scalebox{0.8}{
\begin{tabular}{|c| c| c| c| c| c| c| c| c |c| c ||c |} 
\hline
&  1&2&3&4&5&6&7&8&9&10 & Global \\
\hline
L.E. & $4.65E-2$ & $6.64E-2$ & $4.26E-2$ & $3.11E-2$ & $5.02E-2$ & $1.38E-3$ & $2.19E-2$ & $5.63E-1$ & $5.59E-1$ & $1.99E-2$ & $4.76E-1$ \\
\hline
S.E. & $8.91E-2$ & $3.13E-2$ & $4.75E-2$ & $3.01E-2$ & $1.05E-1$ & $2.99E-3$ & $1.07E-1$ & $6.75E-2$ & $1.03E-1$ & $3.85E-2$ & $1.10E-1$ \\
\hline
Lig. & $1.12E-5$ & $2.62E-5$ & $1.34E-5$ & $6.19E-5$ & $8.35E-5$ & $4.16E-9$ & $5.60E-5$ & $5.07E-5$ & $0$ & $0$ & $6.30E-5$ \\
\hline
Def. & $2.45E-5$ & $8.15E-5$ & $1.40E-4$ & $1.23E-4$ & $1.14E-4$ & $8.80E-9$ & $5.26E-5$ & $5.38E-5$ & $0$ & $0$ & $9.78E-5$ \\
\hline
F.M .& $1.12E-5$ & $2.62E-5$ & $1.37E-5$ & $8.06E-5$ & $1.01E-4$ & $1.38E-9$ & $5.85E-5$ & $5.55E-5$ & $0$ & $0$ & $7.35E-5$ \\
\hline
A.M. & $5.53E-4$ & $4.40E-5$ & $3.63E-4$ & $2.54E-4$ & $1.31E-4$ & $8.84E-8$ & $5.79E-4$ & $1.99E-4$ & $0$ & $0$ & $5.81E-4$ \\
\hline
Food & $2.32E-5$ & $5.59E-5$ & $4.76E-5$ & $3.79E-5$ & $6.72E-5$ & $3.67E-8$ & $2.31E-4$ & $9.71E-5$ & $0$ & $0$ & $1.05E-4$ \\
\hline
RtE & $1.05E+1$ & $8.90$ & $7.76$ & $1.03E+1$ & $1.14E+1$ & $3.82E+1$ & $5.94$ & $1.58E+1$ & $1.77E+1$ & $1.42$ & $1.96E+1$ \\
\hline
B.T.C & $7.46$ & $4.92$ & $6.20$ & $4.20$ & $1.44E+1$ & $3.73E+1$ & $1.12E+1$ & $1.62E+1$ & $1.67E+1$ & $4.12$ & $1.73E+1$ \\
\hline
Den. & $4.53E-3$ & $5.27E-3$ & $4.54E-3$ & $3.42E-3$ & $4.38E-3$ & $9.61E-6$ & $3.06E-3$ & $1.59E-2$ & $1.72E-2$ & $5.48E-3$ & $1.48E-2$ \\
\hline
P.S. & $1.52$ & $1.22$ & $8.62E-1$ & $9.62E-1$ & $9.10E-1$ & $1.03$ & $1.56$ & $3.30$ & $3.68$ & $3.92E-1$ & $3.43$\\
\hline
Lin.X. & $1.04E-2$ & $1.78E-2$ & $4.44E-3$ & $5.63E-3$ & $1.04E-2$ & $1.06E-4$ & $3.69E-3$ & $5.21E-3$ & $5.00E-3$ & $7.07E-3$ & $9.36E-3$\\
\hline
Lin.Y. & $1.01E-2$ & $9.32E-3$ & $1.46E-2$ & $9.55E-3$ & $1.14E-2$ & $7.50E-5$ & $2.44E-3$ & $5.11E-3$ & $4.43E-3$ & $0$ & $9.04E-3$\\
\hline
F.R. & $3.61E-2$ & $6.03E-2$ & $5.55E-2$ & $4.75E-2$ & $5.30E-2$ & $1.60E-3$ & $2.25E-2$ & $7.83E-1$ & $7.90E-1$ & $1.14E-2$ & $6.69E-1$\\
\hline
\end{tabular}
}
\caption{Standard deviation of each generator per metrics, top row abbreviated. Corresponding generator as top row, see list in Appendix A for details.}
\label{Table:6}
\end{table*}
\end{center}

\end{document}